\documentclass[letterpaper]{article} 
\usepackage[]{aaai23}  
\usepackage{times}  
\usepackage{helvet}  
\usepackage{courier}  
\usepackage[hyphens]{url}  
\usepackage{graphicx} 
\usepackage{natbib}  
\usepackage{caption} 
\usepackage{algorithmic}

\usepackage{newfloat}
\usepackage{listings}
\DeclareCaptionStyle{ruled}{labelfont=normalfont,labelsep=colon,strut=off} 
\usepackage{booktabs}       
\usepackage{amsfonts}       
\usepackage{nicefrac}       
\usepackage{microtype}      
\usepackage{amsmath}
\usepackage{xcolor}
\usepackage{multirow}
\usepackage{dsfont}
\usepackage{makecell}
\usepackage{subfig}

\usepackage[ruled,vlined]{algorithm2e}
\SetKwInput{KwInput}{Input}                
\SetKwInput{KwOutput}{Output}              
\SetKw{return}{return}              
\title{Dynamic Structure Pruning for Compressing CNNs}
\author{
    Jun-Hyung Park\textsuperscript{\rm 1},
    Yeachan Kim\textsuperscript{\rm 2},
    Junho Kim\textsuperscript{\rm 2},
    Joon-Young Choi\textsuperscript{\rm 2},
    SangKeun Lee\textsuperscript{\rm 1,2}
}
\affiliations{
    \textsuperscript{\rm 1}Department of Computer Science and Engineering\\
    \textsuperscript{\rm 2}Department of Artificial Intelligence\\
    Korea University, Seoul, Republic of Korea\\

    \{irish07, yeachan, monocrat, johnjames, yalphy\}@korea.ac.kr
}

\begin{document}

\maketitle

\begin{abstract}
Structure pruning is an effective method to compress and accelerate neural networks. While filter and channel pruning are preferable to other structure pruning methods in terms of realistic acceleration and hardware compatibility, pruning methods with a finer granularity, such as intra-channel pruning, are expected to be capable of yielding more compact and computationally efficient networks. Typical intra-channel pruning methods utilize a static and hand-crafted pruning granularity due to a large search space, which leaves room for improvement in their pruning performance. In this work, we introduce a novel structure pruning method, termed as \textit{dynamic structure pruning}, to identify optimal pruning granularities for intra-channel pruning. In contrast to existing intra-channel pruning methods, the proposed method automatically optimizes dynamic pruning granularities in each layer while training deep neural networks. To achieve this, we propose a differentiable group learning method designed to efficiently learn a pruning granularity based on gradient-based learning of filter groups. The experimental results show that dynamic structure pruning achieves state-of-the-art pruning performance and better realistic acceleration on a GPU compared with channel pruning. In particular, it reduces the FLOPs of ResNet50 by 71.85\% without accuracy degradation on the ImageNet dataset. Our code is available at \url{https://github.com/irishev/DSP}.
\end{abstract}

\section{Introduction}
Network pruning removes redundant weights or neurons from neural networks \citep{han2015learning,li2016pruning,he2017channel}. Since deep neural networks (DNNs) are typically over-parameterized \citep{denton2014exploiting,ba2014deep}, pruning techniques can reduce the network size without a significant accuracy loss by identifying and removing redundant weights. Hence, network pruning can address problems in the deployment of DNNs on low-end devices with limited space and computational capacity.

Most approaches in network pruning are classified into two categories; weight pruning \citep{han2015learning,guo2016dynamic} and structure pruning \citep{li2016pruning,he2017channel,mao2017exploring}. Weight pruning deletes individual weights in a network. However, because of the irregular structures generated by weight pruning, it is difficult to leverage the high efficiency of pruned models without using special hardware or libraries \citep{li2016pruning}. In contrast, structure pruning directly discards high-level structures of tensors. Among the structure pruning methods, pruning channels, filters, and layers are commonly more favorable because they can easily achieve acceleration on general hardware and libraries \citep{li2016pruning,he2017channel,he2018soft}. 

Nevertheless, pruning with a finer granularity is expected to be capable of better efficiency, due to the higher degree of freedom of connectivity \citep{mao2017exploring}. Intra-channel pruning \citep{wen2016learning,yu2017accelerating,meng2020pruning} focuses on finding a new pruning granularity that can provide realistic compression and acceleration benefits by decomposing a filter or channel into smaller components. For example, grouped kernel pruning \citep{zhong2021revisit} has found a generally accelerable granularity by grouping filters and pruning channels within a group. Although these methods have empirically shown better pruning rates compared with channel and filter pruning, they have several limitations that can potentially degrade pruning results. First, they typically use a static and strictly constrained pruning granularity in a layer, such as a regular structure of adjacent kernels \cite{wen2016learning} and a channel of evenly grouped filters \cite{zhong2021revisit}. Such granularity limits the degree of freedom, which possibly degrades pruning performance. Second, they utilize hand-crafted heuristics in determining a pruning granularity. In prior methods, a pruning granularity is determined heuristically based on the number of filters \cite{zhong2021revisit} or the shape of kernels \cite{meng2020pruning}. Hence, information on how a set of weights delivers important features for a given task may not be properly identified and utilized. 

\begin{figure*}[t]
  \centering
  \includegraphics[trim={0.5cm 0 0 0},clip,width=0.9\linewidth]{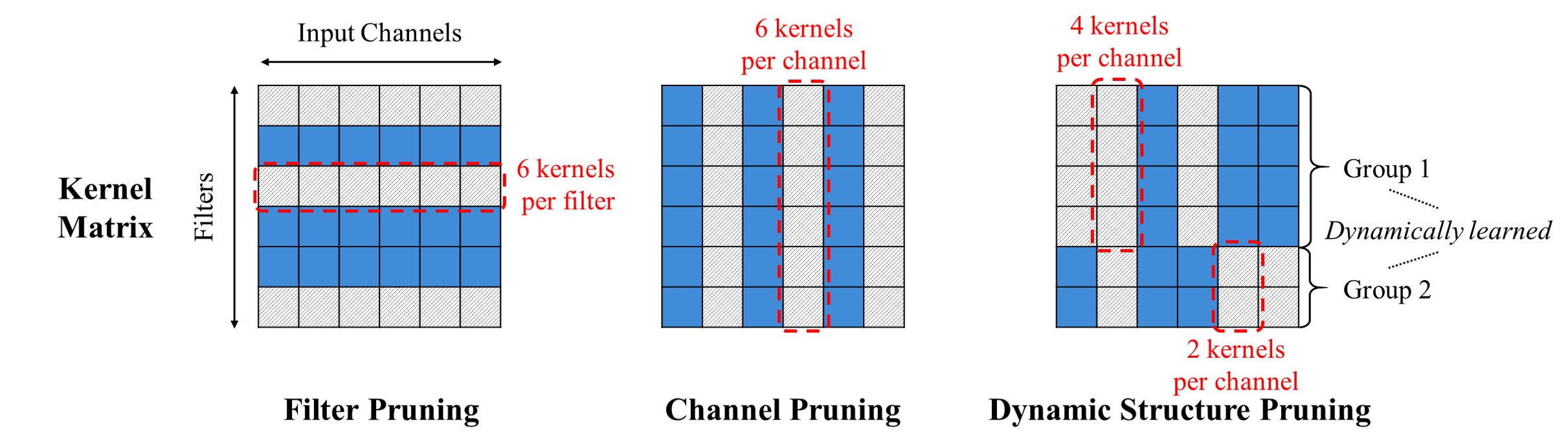}
  \caption{Comparison of pruning granularity among filter, channel, and our dynamic structure pruning.}
  \label{fig:illustration}
\end{figure*}

In this work, we introduce a novel intra-channel pruning method called \textit{dynamic structure pruning} to address the abovementioned issues. Our method automatically learns dynamic pruning granularities in each layer based on the grouped kernel pruning scheme \cite{zhong2021revisit}. As shown in Figure 1, the pruning granularity is changed dynamically depending on which and how many filters are grouped together. Hence, we aim to identify optimal pruning granularities by learning filter groups for effective pruning. Because identifying optimal filter groups is challenging due to their discrete nature and exponential search space, we propose a differentiable group learning method inspired by research on architecture search \cite{liu2018darts,xie2018snas}. We relax the search space to be continuous and approximate the gradient of filter groups with respect to the objective function of a training task. This facilitates efficient learning of filter groups during the gradient-based training of DNNs without time-consuming discrete explorations and evaluations. Consequently, dynamic structure pruning provides a higher degree of freedom on the structural sparsity than filter and channel pruning.

We validate the effectiveness of dynamic structure pruning through extensive experiments with diverse network architectures on the CIFAR-10 \citep{cifar10} and ImageNet \citep{deng2009imagenet} datasets. The results reveal that the proposed method outperforms state-of-the-art structure pruning methods in terms of both pruned FLOPs and accuracy. Notably, it reduces the FLOPs of ResNet50 by 71.85\% without accuracy degradation on the ImageNet inference. Additionally, it exhibits a better accuracy-latency tradeoff on general hardware and libraries than the baselines.

To the best of our knowledge, this work is the first that explores dynamic pruning granularities for intra-channel pruning and provides deep insights. By automatically learning effective pruning granularities, dynamic structure pruning significantly improves pruning results. In addition, it is readily applicable to convolutional and fully-connected layers and generates efficient structured networks, which ensures wide compatibility with modern hardware, libraries, and neural network architectures. The main contributions of this work are summarized as follows:
\begin{itemize}
\item We introduce a novel intra-channel pruning method, called \textit{dynamic structure pruning}, that automatically learns dynamic pruning granularities for effective pruning.
\item We propose a differentiable group learning method that can efficiently optimize filter groups using gradient-based methods during training. 
\item We verify that dynamic structure pruning establishes new state-of-the-art pruning results while maintaining accuracy across diverse networks on several popular datasets.
\end{itemize}

\section{Related Work}

\subsection{Weight Pruning}

A large subset of modern pruning algorithms has attempted to prune the individual weights of networks. To induce weight-level sparsity in a network, each weight with a magnitude less than a certain threshold is removed \citep{han2015learning,guo2016dynamic}. These approaches inevitably generate unstructured models that require specialized hardware and libraries to realize the compression effect \citep{he2018soft}.

\subsection{Structure Pruning} 

\subsubsection{Channel/filter pruning.}
Pruning a network into a regular structure is particularly appealing as a hardware-friendly approach. Typical structure pruning methods \citep{li2016pruning,he2017channel,he2018soft} prune channels and filters, because doing so reduces memory footprints and inference time on general hardware. Data-free methods \citep{li2016pruning,he2018soft,lin2020dynamic} mostly use the $\ell_{p}$-norm of weights to evaluate the importance of a structure. Data-dependent methods empirically evaluate importance based on gradients or activations by utilizing training data \citep{molchanov2019importance,gao2019dynamic,lin2020dynamic}. Several structure pruning methods involve different methods for evaluating the importance of structures, such as their distance from the geometrical median \citep{he2019filter}. Another branch of structure pruning directly induces sparsity in networks during training by utilizing regularization techniques \citep{wen2016learning,yang2019deephoyer,wang2019structured,zhuang2020neuron,wang2020neural}. Moreover, machine learning can be applied to generate compact networks \citep{he2018amc,liu2019metapruning,ye2020good,gao2021network}. These studies have focused predominantly on reconsidering the importance criteria for finding redundant filters and channels more accurately. In contrast, our method considers a finer granularity than a filter or channel to generate more compact and easily accelerable networks. 

\subsubsection{Intra-channel pruning.}
Another line of work exploits an intra-channel-level sparsity to achieve better efficiency from a finer granularity. \citet{wen2016learning} introduce group-wise pruning, which induces shape-level structured sparsity in neural networks. \citet{mao2017exploring} have further explored a wide range of pruning granularity and evaluated its effects on prediction accuracy and efficiency. \citet{meng2020pruning} have proposed stripe-wise pruning to learn the shape of filters. In most cases, an intra-channel-level sparsity requires a specialized library to realize its efficiency, which limits its applicability. Grouped kernel pruning \citep{yu2017accelerating,zhang2022group,zhong2021revisit} has attracted attention due to its compatibility. These methods prune a constant number of channels within evenly grouped filters and consequently yield pruned networks that can be executed using a grouped convolution, which is widely supported by modern hardware and libraries. Intra-channel pruning methods typically leverage a static and hand-crafted pruning granularity with strict constraints. In contrast, we explore a dynamic pruning granularity based on gradient-based learning to maximize efficiency with an intra-channel-level sparsity.

\subsubsection{Other approaches.} Other structure pruning approaches combine structure pruning with different network compression methods such as matrix decomposition \citep{li2020group}, factorization \citep{li2019compressing}, and skip calculation \citep{tang2021manifold}, which are distantly related to the proposed method.

\begin{figure*}[t]
  \centering
  \includegraphics[width=0.9\linewidth]{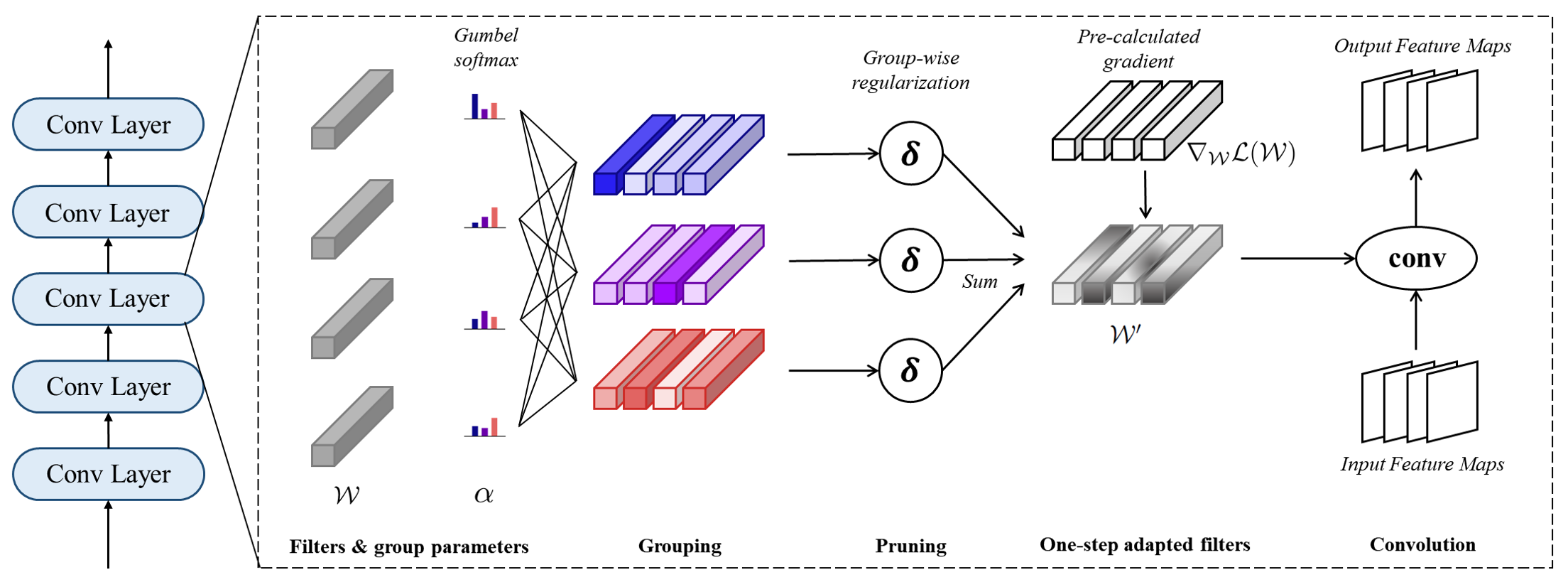}
  \caption{Illustration of obtaining one-step adapted filters $\mathcal{W}'$ for differentiable group learning.}
  \label{fig:gap}
\end{figure*}

\section{Dynamic Structure Pruning}

In this section, we formulate dynamic structure pruning and propose a differentiable group learning method to optimize filter groups. Then, we illustrate our group channel pruning method. The entire process is described in Algorithm 1.

\subsection{Problem Formulation}

We provide a formal description of dynamic structure pruning. Given a weight tensor of a convolutional layer, we aim to search for $N$ filter groups. Let $L$ be the number of convolutional layers, and the weight tensor of $i$-th convolutional layer can be represented by $\mathcal{W}^{i} \in \mathbb{R}^{C^{i}_{out} \times C^{i}_{in} \times K_h^i \times K_w^i}$, $i=1,...,L$, where $C^{i}_{out}$, $C^{i}_{in}$, $K_h^i$, and $K_w^i$ are the number of output channels (i.e., the number of filters), input channels, kernel height, and kernel width, respectively. Then, a filter group of the layer is defined as:
\begin{equation}
\begin{aligned}
\mathcal{G}^{i}_{p} = \{\mathcal{W}^{i}_{k} | \alpha^{i}_{k,p}=1, k \in \{1,...,C^{i}_{out}\}\} \qquad\\
\text{s.t. } \, p \in \{1,...,N\} , \, \cup_{q=1}^{N}\mathcal{G}^{i}_{q} = \mathcal{G}^{i}, \qquad \quad \\
(\forall m, n \in \{1,...,N\}) \, m \neq n \Rightarrow \mathcal{G}^{i}_{m} \cap \mathcal{G}^{i}_{n} = \emptyset,
\end{aligned}
\end{equation}
where $\alpha^{i}_{k,p} \in \{0, 1\}$ is a binary group parameter, which indicates whether $k$-th filter $\mathcal{W}^i_k$ belongs to $p$-th group $\mathcal{G}^i_p$; $N$ is the pre-defined number of groups; and $\mathcal{G}^{i}=\{\mathcal{W}^{i}_{1}, . . . , \mathcal{W}^{i}_{C^{i}_{out}}\}$ is the set of all filters of the $i$-th layer. Following our definition, all filters belong to one and only one group. Subsequently, we separately prune input channels of each filter group, termed as \textit{group channels}; essentially, this implies that filter groups determine pruning granularities. Therefore, we optimize the set of all group parameters $\mathbf{\alpha}$, which determine filter groups, along with the set of all weight tensors $\mathcal{W}$ to minimize the training loss. Let $\mathcal{L}(\cdot)$ and $\delta(\cdot)$ denote the training loss function and the pruning operator, respectively. We solve a bilevel optimization problem with $\alpha$ as the upper-level variable and $\mathcal{W}$ as the lower-level variable:
\begin{equation}
\begin{aligned}
\min_{\mathbf{\alpha}} \mathcal{L}(\delta(\mathcal{W^*}(\mathbf{\alpha}),\alpha)) \qquad\qquad \\
\text{s.t. } \, \mathcal{W^*}(\mathbf{\alpha})={\arg\min}_{\mathcal{W}} \, \mathcal{L}(\delta(\mathcal{W}, \alpha)).
\end{aligned}
\end{equation}
This formulation is also found in the work of differentiable architecture search \cite{liu2018darts}, which can be explained in that filter groups can be considered as a special type of architecture.

\begin{algorithm}[t]
\small
\SetAlgoLined
 \tcp{1. Differentiable Group Learning}
 \While{not converged}{
  $\alpha \leftarrow \alpha - \epsilon_1 \nabla_{\mathcal{\alpha}} \mathcal{L}(\delta(\mathcal{W^*}(\mathbf{\alpha}),\alpha))$\\
  $\mathcal{W} \leftarrow \mathcal{W} - \epsilon_2 \nabla_{\mathcal{W}} \mathcal{L}(\delta(\mathcal{W^*}(\mathbf{\alpha}),\alpha))$
 }
 \tcp{2. Group Channel Pruning}
 \For{$i=1$ to $L$}{
  $\alpha^i \leftarrow \text{Discretize}(\alpha^i)$\\
  \For{$p=1$ to $N$}{
   $\mathcal{I}^i_{p} \leftarrow \{||\alpha^i_{:,p}\mathcal{W}^i_{:,1,:,:}||^2_2$, ... , $||\alpha^i_{:,p} \mathcal{W}^i_{:,C^i_{in},:,:}||^2_2$\}\\
   $C^i_p \leftarrow \text{FindRedundantChannel}(\mathcal{I}^i_{p}, \mathcal{W}^i)$\\
  }
  $\mathcal{W}^i_{prune} \leftarrow \text{PruneGroupChannel}(\mathcal{W}^i, \alpha^i, C^i)$
 }
 \tcp{3. Fine-tuning}
 \While{not converged}{
  $\mathcal{W}_{prune} \leftarrow \mathcal{W}_{prune} - \epsilon \nabla_{\mathcal{W}_{prune}}\mathcal{L}(\mathcal{W}_{prune})$\\
 }
 \return{$W_{prune}$}
 \caption{\textsc{Dynamic Structure Pruning}}
\end{algorithm}

\subsection{Differentiable Group Learning}

Next, we propose a novel differentiable method to jointly learn group parameters $\alpha$ and weights $\mathcal{W}$. The optimization problem in Eq. 2 has non-differentiable factors, which prevent the application of standard gradient-based methods, such as back-propagation and stochastic gradient descent. Therefore, we introduce several approximation techniques to facilitate gradient calculation.

\subsubsection{Continuous relaxation of $\alpha$.} Because $\alpha$ is a discrete variable, the back-propagation algorithm cannot be applied to compute its gradients. Hence, we use a gumbel-softmax reparameterization trick \cite{jang2016categorical} to perform continuous relaxation of $\alpha$. We generate $\alpha^i_{k,p} \in \alpha$ from learnable parameters $\pi^i_{k,1},...,\pi^i_{k,N}$ as follows:
\begin{equation}
\begin{aligned}
\alpha^i_{k,p} = \frac{\exp{((\log(\pi^i_{k,p})+g^i_{k,p})/\tau)}}{\sum^N_{j=1}\exp((\log(\pi^i_{k,j})+g^i_{k,j})/\tau)},
\end{aligned}
\end{equation}
where $g^i_{k,1},...,g^i_{k,N}$ are i.i.d samples drawn from a Gumbel distribution $\texttt{Gumbel}(0,1)$, and $\tau$ is a temperature parameter to control sharpness. When we need discrete values after training, we calculate a one-hot vector $a^i_k$ with $a^i_{k,p}=1$, where $p=\arg\max_{j=1, ... , N} \pi^i_{k,j}$.

\subsubsection{Regularization-based pruning.} Among various approaches for a pruning operator $\delta(\cdot)$, we use a regularization-based approach with a group-wise regularizer $R(\cdot)$ due to its differentiability and smoothness as follows:
\begin{equation}
\begin{aligned}
\min_{\mathbf{\alpha}} (\mathcal{L}(\mathcal{W^*}(\alpha))  + \lambda R(\mathcal{W^*}(\alpha), \mathbf{\alpha})), \qquad \quad \\
\text{s.t. } \, \mathcal{W^*}(\mathbf{\alpha})={\arg\min}_{\mathcal{W}} \, (\mathcal{L}(\mathcal{W}) + \lambda R(\mathcal{W}, \mathbf{\alpha})), \quad \\
R(\mathcal{W}, \mathbf{\alpha}) = \sum_{i \in I} \sum_{p=1}^{N} \sum_{m=1}^{C^{i}_{in}} |\alpha^i_{:,p}|_{0.5} \sqrt{\sum_{k=1}^{C^i_{out}}||\alpha^i_{k,p} W^{i}_{k,m}||_{2}^{2}},
\end{aligned}
\end{equation}
where $\lambda$ is a regularization hyper-parameter and I is the set of indices of layers in which the filters are grouped. This regularizer induces sparsity at group channels of each $W^{i}$. Note that we adaptively scale $\lambda$ for each layer. 

\subsubsection{One-step unrolling.} In Eq. 4, calculating the gradient of $\alpha$ can be prohibitive due to the costly optimization of $\mathcal{W^*}(\alpha)$. We approximate $\mathcal{W^*}(\alpha)$ by using filters adapted with a single training step as follows:
\begin{equation}
\begin{aligned}
\mathcal{W}' = \mathcal{W} - \epsilon \nabla_{\mathcal{W}}\mathcal{L}(\mathcal{W}) - \epsilon\lambda\nabla_{\mathcal{W}} R(\mathcal{W}, \mathbf{\alpha}),
\end{aligned}
\end{equation}
where $\epsilon$ is a learning rate.  The one-step adapted filters $\mathcal{W'}$ can be obtained using group-wise regularization and pre-calculated gradient of filters, as shown in Figure 2. From Eq. 4 and 5, we derive the gradient by applying the chain rule:
\begin{equation}
\begin{aligned}
&\lambda\nabla_{\mathbf{\alpha}} R(\mathcal{W}', \mathbf{\alpha}) \\ -&\epsilon\lambda \nabla_{\mathbf{\alpha},\mathcal{W}}^2 R(\mathcal{W}, \mathbf{\alpha}) \nabla_{\mathcal{W}'} (\mathcal{L}(\mathcal{W}')+ \lambda R(\mathcal{W'}, \mathbf{\alpha})),
\end{aligned}
\end{equation}
Note that we assume that $\alpha^i_{k,p}$ is independent of variables other than $W^i_{k}$ for efficiency, because this assumption enables us to analytically calculate the second derivative from Eq. 4.

\subsection{Group Channel Pruning}

After optimizing group parameters, we prune group channels. The majority of existing pruning methods \cite{you2019gate,molchanov2019importance} conduct iterative optimization procedures, which require numerous evaluations during prune-retrain cycles. These incur a considerable computational cost, particularly when pruning at an extreme sparsity level. Therefore, we introduce a simple yet effective one-shot pruning method. Given a set of weights $\mathcal{S}$, we adaptively prune each group with the maximum pruning rate that satisfies the following condition:
\begin{equation}
\begin{aligned}
\frac{||Q^i_p||_2^2}{||\mathcal{G}^i_p||_2^2} < \beta,
\end{aligned}
\end{equation}
where $Q^i_p$ is the weights connected to pruned channels of $p$-th group in the $i$-th layer, and $\beta$ is a pre-defined hyper-parameter. Intuitively, this constrains the relative amount of the information to be removed from group channels.

\begin{table*}[t]
\caption{Pruning results on CIFAR-10. Our baseline accuracy is reported in parentheses below the network name.}
\centering
\small
\begin{tabular}{clcccc}
\toprule
Network & Method & \makecell{P. Acc. (\%)} & \makecell{$\Delta$Acc. (\%)} & \makecell{Params$\downarrow$ (\%)} & \makecell{FLOPs$\downarrow$ (\%)}\\ \midrule
\multirow{6}{*}{\makecell{ResNet20\\(91.90\%)}}
&TMI-GKP \citep{zhong2021revisit}&  92.01 & -0.34 & 43.35 & 42.87 \\
&Rethinking \citep{ye2018rethinking} & 90.90 & -1.34 & 37.22 & 47.40 \\
&DHP \citep{li2020dhp} & 91.53 & -1.00 & 43.87 & 48.20 \\
&DSP (g=2) (ours)& 91.91 & +0.01 & 50.67 & 55.76 \\
&DSP (g=3) (ours)& 92.00 & +0.10 & 53.22 & 57.91 \\
&DSP (g=4) (ours)& \textbf{92.16} & \textbf{+0.26} & \textbf{53.86} & \textbf{58.81} \\\midrule 
\multirow{13}{*}{\makecell{ResNet56\\(93.26\%)}}
&HRank \citep{lin2020hrank}&  93.52 & +0.26 & 16.80 & 29.30\\
&NISP \citep{yu2018nisp}&  93.01 & -0.03 & 42.40 & 35.50\\
&GAL \citep{lin2019towards}&  93.38 & +0.22 & 11.80 & 37.60\\
&TMI-GKP \citep{zhong2021revisit}&  94.00 & +0.22 & 43.49 & 43.23 \\
&CHIP \citep{sui2021chip}&  94.16 & +0.90 & 42.80 & 47.40\\
&KSE \citep{li2019exploiting}&  93.23 & +0.20 & 45.27 & 48.00\\\
&Random \citep{li2022revisiting}&  93.48 & +0.22 & 44.92 & 48.97 \\
&DHP \citep{li2020dhp}&  93.58 & +0.63 & 41.58 & 49.04 \\
&TDP \citep{wang2021accelerate}&  93.76 & +0.07 & 40.00 & 50.00 \\
&Hinge \citep{li2020group}&  93.69 & +0.74 & 48.73 & 50.00\\
&DSP (g=2) (ours)&  93.92 & +0.66 & 57.33 & 58.56  \\
&DSP (g=3) (ours)&  94.08 & +0.82 & 58.06 & 59.89  \\
&DSP (g=4) (ours)&  \textbf{94.19} & \textbf{+0.93} & \textbf{59.50} & \textbf{60.56}  \\ \midrule
\multirow{7}{*}{\makecell{VGG16\\(93.88\%)}}
&GAL \citep{lin2019towards}&  93.77 & -0.19 & 77.60 & 39.60\\
&HRank \citep{lin2020hrank}&  93.43 & -0.53 & \textbf{82.90} & 53.50\\
&CHIP \citep{sui2021chip}&  93.86 & -0.10 & 81.60 & 58.10\\
&Hinge \citep{li2020group}&  93.59 & -0.43 & 19.95 & 60.93\\
&DSP (g=2) (ours) & 93.88 & +0.00 & 74.51 & 75.58 \\
&DSP (g=3) (ours) & \textbf{93.91} & \textbf{+0.03} & 76.65 & 77.80 \\
&DSP (g=4) (ours) & \textbf{93.91} & \textbf{+0.03} & 76.93 & \textbf{80.51} \\\bottomrule
\end{tabular}
\end{table*}

\begin{table*}[t]
\caption{Pruning results on ImageNet. Our baseline top-1 / top-5 accuracy is reported in parentheses below the network name.}
\small
\centering
\begin{tabular}{@{}c@{\quad}l@{}c@{\quad}c@{\quad}c@{\quad}c@{\quad}c@{\quad}c@{}}
\toprule
Network & Method & \makecell{Top-1\\P. Acc. (\%)} & \makecell{Top-1\\$\Delta$Acc. (\%)} & \makecell{Top-5\\P. Acc. (\%)}   & \makecell{Top-5\\$\Delta$Acc. (\%)}& \makecell{Params$\downarrow$ (\%)}& \makecell{FLOPs$\downarrow$ (\%)}\\ \midrule
\multirow{5}{*}{\makecell{ResNet18\\(69.76\% / 89.08\%)}} 
&PFP \citep{liebenwein2019provable} &  67.38  & -2.36& 87.91 & -1.16 & 43.80 & 29.30  \\
&SCOP \citep{tang2020scop} &  68.62 & -1.14 & 88.45 & -0.63 & 43.50 & 45.00  \\
&DSP (g=2) (ours)&  69.28 & -0.48 & 88.59 & -0.49 & 54.14  & 60.43  \\
&DSP (g=3) (ours)&  69.30 & -0.46 & 88.61 & -0.47 & 54.87 & 61.21  \\
&DSP (g=4) (ours)&  \textbf{69.38} & \textbf{-0.38} & \textbf{88.77} & \textbf{-0.31} & \textbf{54.99}  & \textbf{61.33}  \\\midrule
\multirow{12}{*}{\makecell{ResNet50\\(76.13\% / 92.86\%)}}
&TMI-GKP \citep{zhong2021revisit} &  75.33 & -0.62 & - & - & 33.21 & 33.74  \\
&ThiNet \citep{luo2017thinet} &  72.04 & -0.84 & 91.14 & -0.47 & 33.70 & 36.80  \\
&GAL \citep{lin2020hrank} &  71.95 & -4.20 & 90.94 & -1.93 & 16.90 & 43.00 \\
&HRank \citep{lin2020hrank} &  74.98 & -1.17 & 92.33 & -0.54 & 36.60 & 43.70 \\
&SCOP \citep{tang2020scop} &  75.95 & -0.20 & 92.79 & -0.08 & 42.80 & 45.30 \\
&CHIP \citep{sui2021chip} &  76.15 & +0.00 & 92.91 & +0.04 & 44.20 & 48.70 \\
&TDP \citep{wang2021accelerate} &  75.90 & -0.25 & - & - & 53.00 & 50.00 \\
&Random \citep{li2022revisiting} &  75.13 & -0.20 & 92.52 & -0.08 & 45.88 & 51.01 \\
&DCP \citep{zhuang2018discrimination} &  74.95 & -1.06 & 92.32 & -0.61 & 51.56 & 55.50 \\
&DSP (g=2) (ours)& 76.16 & +0.03 & 92.89 & +0.03 & 60.88 & 67.78 \\
&DSP (g=3) (ours)& 76.20 & +0.07 & 92.90 & +0.04 & 62.23 & 70.98 \\
&DSP (g=4) (ours)& \textbf{76.22} & \textbf{+0.09} & \textbf{92.92} & \textbf{+0.06} & \textbf{63.45} & \textbf{71.85} \\\midrule
\multirow{6}{*}{\makecell{MobileNetV2\\(71.88\% / 90.29\%)}} 
&MP \citep{liu2019metapruning} & 71.20 & +0.00 & - & - & - & 27.67 \\
&Random \citep{li2022revisiting} &  70.90 & -0.30 & - & - & - & 29.13 \\
&AMC \citep{he2018amc} & 70.80 & -0.40 & - & - & - & 30.00 \\
&DSP (g=2) (ours)& 71.60 & -0.28 & 90.14 & -0.15 & 31.80 & 37.01  \\
&DSP (g=3) (ours)& 71.60 & -0.28 & 90.15 & -0.14 & 31.87 & 37.10 \\
&DSP (g=4) (ours)& \textbf{71.61} & \textbf{-0.27} & \textbf{90.17} & \textbf{-0.12} & \textbf{31.95} & \textbf{37.24}  \\
\bottomrule
\end{tabular}
\end{table*}

\section{Experiments}

In this section, we present experimental results that demonstrate the efficacy of dynamic structure pruning. We evaluate dynamic structure pruning on two image classification benchmarks including CIFAR-10 \citep{cifar10} and ImageNet \citep{deng2009imagenet} using the prevalent ResNet \citep{he2016deep} following the related work \citep{he2018soft}. In addition, to further evaluate its applicability, we extend our experiments to VGG \citep{simonyan2014very} and MobileNetV2 \citep{sandler2018mobilenetv2}. We primarily compare our proposed method to state-of-the-art channel and filter pruning methods. In addition, we compare our method to state-of-the-art intra-channel pruning methods such as TMI-GKP \cite{zhong2021revisit}. We evaluate dynamic structure pruning by setting the number of groups to two, three, and four, which is denoted in form of ``DSP (g=\{number of groups\})''. Note that we omit the results of more than four groups because we have not observed any significant differences from those of four groups. We report test/validation accuracy of pruned models (P. Acc.) for CIFAR-10/ImageNet, accuracy difference between the original and pruned models ($\Delta$Acc.), and pruning rates of parameters (Params$\downarrow$) and FLOPs (FLOPs$\downarrow$). We report the average results of five runs on a single pre-trained model.

\subsubsection{Experimental Settings.} In the CIFAR-10 experiments, we manually pre-train the original networks using the standard pre-processing and training settings in \citet{he2016deep}. In the ImageNet experiments, we use the pre-trained checkpoints provided by Pytorch\footnote{\url{https://github.com/pytorch/examples/tree/master/imagenet}}  \citep{pytorch}. We search the hyperparameters for dynamic structure pruning based on the empirical analysis, i.e., the value of $\tau \in \{0.125, 0.25, 0.5, 1\}$, $\lambda \in \{$5e-4, 1e-3, 2e-3, 3e-3$\}$ for CIFAR-10 and  $\lambda \in \{$1e-4, 2e-4, 3e-4, 5e-4$\}$ for ImageNet. We use Adam optimizer with a learning rate of 0.001 and momentum of $(0.9, 0.999)$ to train group parameters. During differentiable group learning, we set the initial learning rate to 0.05, and train models for 120 and 60 epochs in the CIFAR-10 and ImageNet experiments, respectively. Then, pruned models are fine-tuned for 80 epochs with initial learning rates of 0.015 and 0.05 for five and three iterations in the CIFAR-10 and ImageNet experiments, respectively. We use a cosine learning rate scheduling with weight decay of 1e-3 and 3e-5 for the CIFAR-10 and ImageNet experiments, respectively, to yield the best results fitted to our additional regularization. Note that we further prune filters of the final convolutional layer in each residual block using $\beta$ to maximize pruning rates. The experiments are implemented using Pytorch and conducted on a Linux machine with an Intel i9-10980XE CPU and 4 NVIDIA RTX A5000 GPUs.

\subsection{Experimental Results}

\subsubsection{Results on CIFAR-10.} We evaluate the pruning performance on the CIFAR-10 dataset with ResNet20, ResNet56, and VGG16. The pruning results shown in Table 1 indicate that dynamic structure pruning generates more compact and computationally efficient models than baseline pruning methods. Without sacrificing accuracy, dynamic structure pruning achieves significantly higher pruning rates than the baseline methods for all the given networks in this experiment. In particular, dynamic structure pruning shows competitive accuracy by using about 75\% of the FLOPs compared with a state-of-the-art pruning method (CHIP) with ResNet56. These results confirm that dynamic structure pruning can effectively learn dynamic pruning granularities and eliminate redundancy on a wide variety of networks. Moreover, dynamic structure pruning achieves greater improvements in pruning results compared with the baselines as the model complexity increases. In addition, we observe that dynamic structure pruning with four groups outperforms that with fewer groups, which shows that more groups tend to be beneficial in terms of accuracy with commensurate pruning rates.

\subsubsection{Results on ImageNet.}

To verify the scalability and real-world applicability of dynamic structure pruning, we perform further evaluation on the ILSVRC-2012 dataset \citep{deng2009imagenet} with ResNet18, ResNet50, and MobileNetV2. The results presented in Table 2 confirm that dynamic structure pruning can consistently generate more efficient and accurate models than the baselines on the large-scale dataset and models. Similar to the results on the CIFAR-10 dataset, dynamic structure pruning reduces more parameters and FLOPs while maintaining the original accuracy, whereas most of the baseline methods show significant accuracy degradation. Particularly, dynamic structure pruning with ResNet50 achieves better accuracy while using about half of the FLOPs, compared with CHIP.

\begin{figure}[t]
\centering
\includegraphics[width = \linewidth]{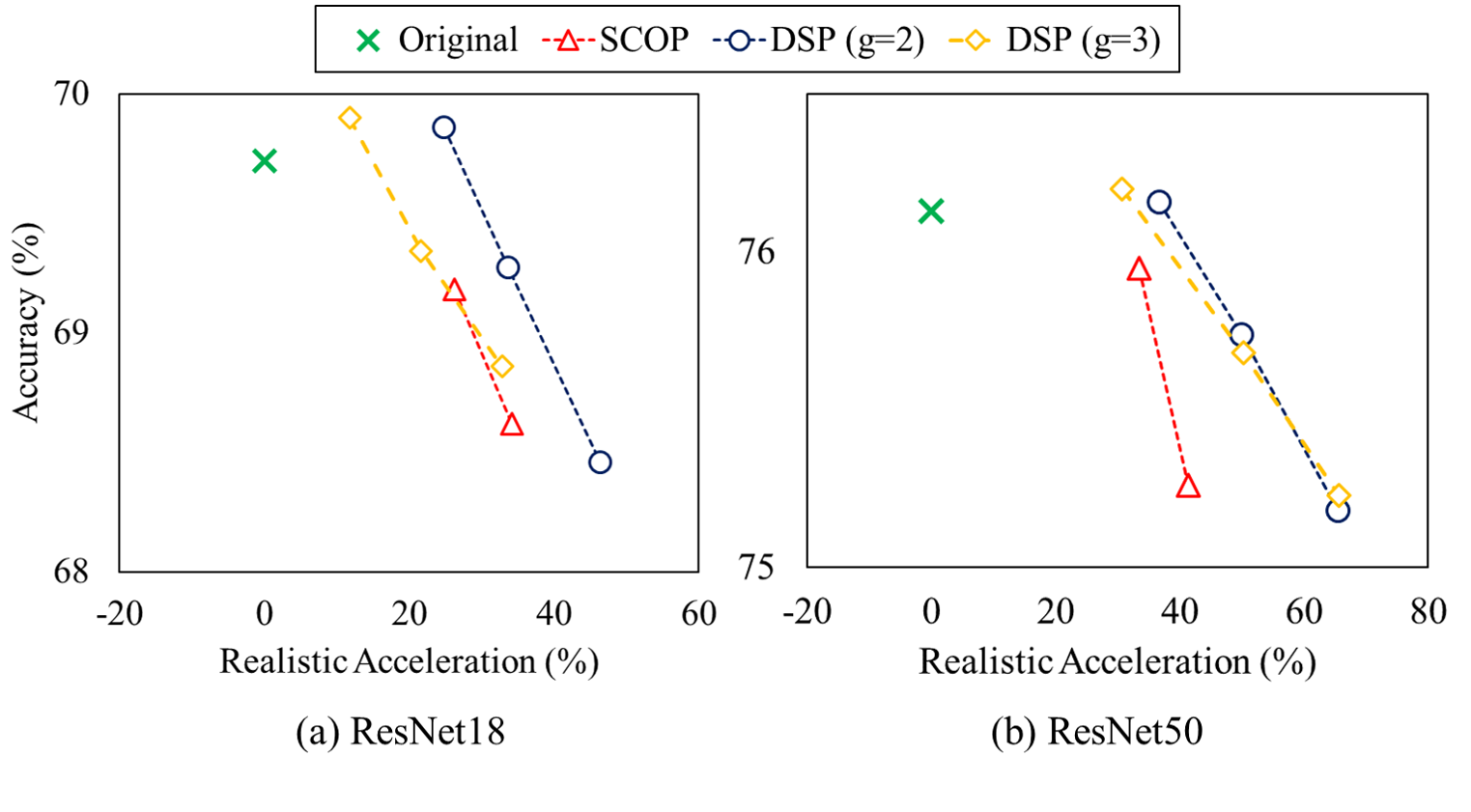}
\caption{ImageNet top-1 accuracy vs. acceleration on a GPU.}
\end{figure}

\subsubsection{Realistic acceleration.}

Because implementing dynamic structure pruning introduces sparse operations of gathering input channels, we consider the possibility of a gap between the theoretical acceleration (i.e. pruned FLOPs) and the realistic acceleration. To evaluate the realistic acceleration in general environments, we measure the realistic acceleration of dynamic structure pruning on the ImageNet inference with a batch size of 128 using an NVIDIA RTX A5000 GPU, and compare it with that of a state-of-the-art channel pruning method (SCOP) that has reported the realistic acceleration. As shown in Figure 3, dynamic structure pruning with two groups achieves a significant improvement in the accuracy-latency tradeoff compared with SCOP. This result indicates that sparsity in dynamic structure pruning can be effectively exploited by general hardware and libraries. In addition, as can be seen in Figure 3, we observe that dynamic structure pruning with three groups typically shows a worse accuracy-latency tradeoff than with two groups despite better pruning performance. This is because the realistic speedup only comes when the reduced computational time outweighs the increased time for gathering channels for additional groups. In this experiment, we find that two groups are generally the best option in terms of an accuracy-latency tradeoff, and more groups may be better with an extreme sparsity level. 

\subsection{Analysis}

\begin{figure*}[t]
\centering
\includegraphics[width = \linewidth]{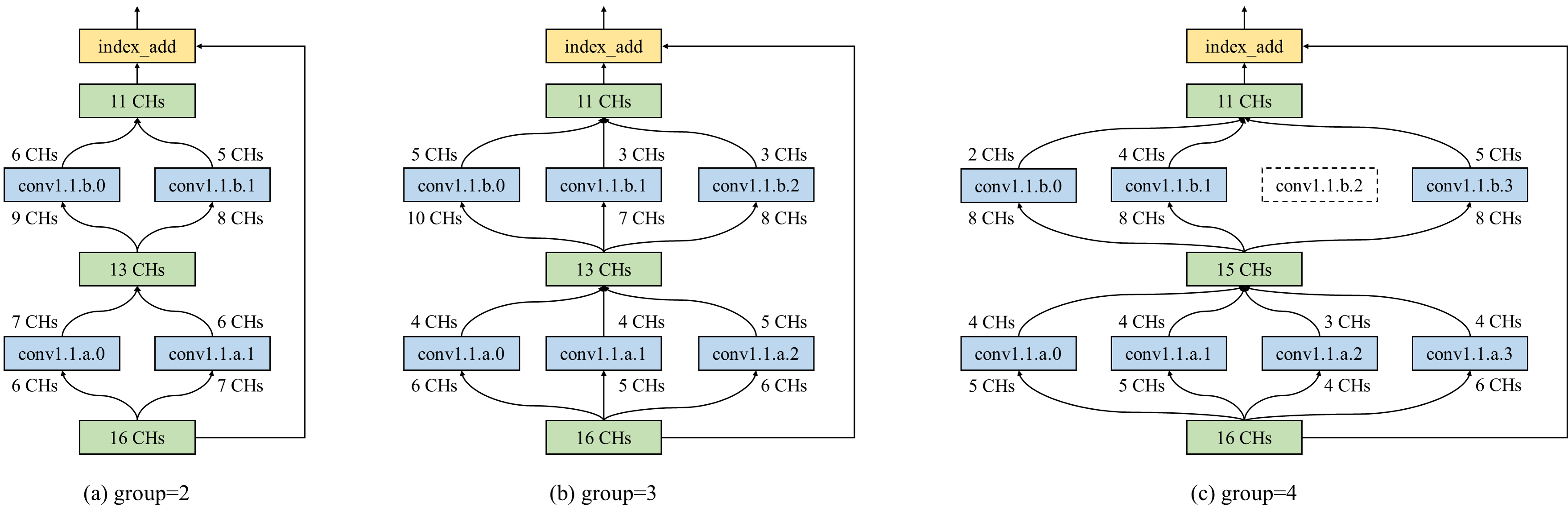}
\caption{Pruned group structure in the ResNet20 block. Dynamic structure pruning allows input channels to be used repeatedly by multiple groups, while output channels are unique for each group.}
\end{figure*}

\begin{table*}[t]
\caption{Comparison of filter groups learned by differentiable group learning with the best, worst, and average cases found by brute force. \underline{Underlined} denotes the second-best accuracy.}
\centering
\small
\begin{tabular}{c|cc|cc|cc|cc}
\toprule
\multirow{2}{*}{Pruning rates} & \multicolumn{2}{c}{Best} &  \multicolumn{2}{c}{Worst} &  \multicolumn{2}{c}{Average} & \multicolumn{2}{c}{Ours} \\
& Groups & Acc. & Groups & Acc. & Groups & Acc. & Groups & Acc.  \\\midrule
0.25 & [0,1,4,5,6] [2,3,7] & \textbf{97.03} & [0,2,4,7] [1,3,5,6] & 59.07 & - & 92.85 & \multirow{3}{*}{[0,1,2,5,6,7] [3,4]} & \underline{96.05} \\
0.5 & [0,1,2] [3,4,5,6,7] & \textbf{88.41} & [0,3,4,7] [1,2,5,6] & 25.74 & - & 66.33 &  & \underline{81.77} \\
0.75 & [0,3,6,7] [1,2,4,5] & \textbf{71.28} & [0,3,4,5,7] [1,2,6] & 10.16 & - & 41.63 &  & \underline{45.12} \\
\bottomrule
\end{tabular}
\end{table*}

\begin{table}[t]
\caption{Compatibility analysis of filter groups in Table 3.}
\centering
\small
\begin{tabular}{lcccc}
\toprule
\multirow{2}{*}{Method} & \multirow{2}{*}{Groups} & \multicolumn{3}{c}{Pruning rates}\\
& & 0.25 & 0.5 & 0.75\\\midrule
\multirow{3}{*}{\makecell{Brute force}} & [0,1,4,5,6] [2,3,7] & \textbf{97.03} & 66.35 & 43.78 \\
& [0,1,2] [3,4,5,6,7] & 95.29 & \textbf{88.41} & 19.97 \\
& [0,3,6,7] [1,2,4,5] & 95.29 & 69.37 & \textbf{71.28}\\ \midrule
Ours & [0,1,2,5,6,7] [3,4] & \underline{96.05} & \underline{81.77} & \underline{45.12}\\
\bottomrule
\end{tabular}
\end{table}

\subsubsection{Pruned group structure.}
To further understand the improved pruning performance, we inspect the grouping and pruning results produced by dynamic structure pruning. Figure 4 illustrates the structure of the second block in ResNet20 pruned by dynamic structure pruning. We find several patterns in the block structure that may be advantageous to pruning performance. First, each group comprises and utilizes different filters and input channels, which suggests that dynamic structure pruning can adaptively identify the number of filters and channels for each group for optimal performance. Second, dynamic structure pruning can adjust the number of groups by pruning redundant filter groups as shown in Figure 4(c). These observations show that dynamic structure pruning indeed provides a higher degree of freedom to pruned structures than standard channel pruning, leading to improvements in pruning performance.

\subsubsection{Empirical analysis of differentiable group learning.}
Although we are currently unaware of theoretical guarantees that our differentiable group learning finds optimal groups, we have empirically observed that it yields acceptable groups that perform well with a wide range of pruning rates. We compare the groups learned by differentiable group learning with the best, worst, and average cases found by brute force. We evaluate the methods on the MNIST dataset \cite{deng2012mnist} with a toy network with two convolutional layers, which output 8 channels, followed by batch normalization and ReLU activation layers. We set the initial learning rate, number of epochs, and $\lambda$ to 0.1, 40, and 1e-3, respectively. We use SGD with the 0.9 momentum factor and exponential learning rate schedule with the 0.9 decay factor. Note that we report the validation accuracy of a pruned network with two groups without fine-tuning. We apply dynamic structure pruning to the second convolutional layer and learn the filter groups from scratch. The results are reported in Tables 3 and 4. Table 3 shows that dynamic group learning effectively finds group composition that exhibits better pruned accuracy than the average case. Our method achieves an accuracy close to the best with low and moderate pruning rates (i.e., 0.25 and 0.5), while still achieving better accuracy than average with a high pruning rate (i.e., 0.75). Though dynamic group learning has not found the best case obtained via brute force in this experiment, it has found groups more compatible with various pruning rates. As shown in Table 4, the best case of the brute force method exhibits groups overfitted to a specific pruning rate, leading to low accuracy with different pruning rates. In contrast, dynamic group learning achieves the second-best accuracy with all pruning rates in this experiment, which confirms its compatibility.

\section{Conclusion}

In this work, we have introduced dynamic structure pruning, which automatically learns dynamic pruning granularities for intra-channel pruning. The results of our empirical evaluations on popular network architectures and datasets demonstrate the efficacy of dynamic structure pruning in obtaining compact and efficient models. In particular, we have found that dynamic structure pruning consistently outperforms the state-of-the-art structure pruning methods in terms of pruned FLOPs, while also exhibiting better accuracy. Moreover, our results reveal that dynamic structure pruning generates more accelerated models on general hardware and libraries than conventional channel pruning. We plan to investigate the efficacy of dynamic structure pruning on other DNN architectures like Transformers \cite{vaswani2017attention} in the future.

\section*{Acknowledgement}
This work was supported by the Basic Research Program through the National Research Foundation of Korea (NRF) grant funded by the Korea government (MSIT) (2020R1A4A1018309), National Research Foundation of Korea (NRF) grant funded by the Korea government (MSIT) (2021R1A2C3010430) and Institute of Information communications Technology Planning Evaluation (IITP) grant funded by the Korea government (MSIT) (No. 2019-0-00079, Artificial Intelligence Graduate School Program (Korea University)).

\bibliography{aaai23}

\end{document}